\documentclass[sigconf]{acmart}
\AtBeginDocument{%
  }

\setcopyright{rightsretained}
\copyrightyear{2024}
\acmYear{2024}
\acmConference[CODS-COMAD Dec '24]{8th International Conference on Data Science and Management of Data (12th ACM IKDD CODS and 30th COMAD)}{December 18--21, 2024}{Jodhpur, India}
\acmBooktitle{8th International Conference on Data Science and Management of Data (12th ACM IKDD CODS and 30th COMAD) (CODS-COMAD Dec '24), December 18--21, 2024, Jodhpur, India}
\acmPrice{}
\acmDOI{10.1145/3703323.3703723}
\acmISBN{979-8-4007-1124-4/24/12}
\usepackage{microtype}
\usepackage{graphicx}
\usepackage{amsmath}
\usepackage{float}
\usepackage{booktabs}
\usepackage{pdfrender}
\usepackage{algorithm}
\usepackage{algpseudocode}
\usepackage{pifont}
\usepackage{wasysym}
\usepackage{enumitem}

\newcommand\doublecheck{{\checked\kern-1em\checked}}
\newcommand{\cross}{
    \textpdfrender{
        TextRenderingMode=Fill,
        LineWidth=0.01pt,
        LineDashPattern=[1 1]0,
    }{\text{\sffamily x}}
}
\title{Enhancing Customer Service Chatbots with Context-Aware NLU through Selective Attention and Multi-task Learning}

\author{Subhadip Nandi}
\email{subhadip.nandi@walmart.com}
\affiliation{%
  \institution{Walmart Inc.}
  \country{Bengaluru, India}
  }

\author{Neeraj Agrawal}
\email{neeraj.agrawal@walmart.com, aneeraj@iisc.ac.in}
\affiliation{%
  \institution{Walmart Inc.}
  \institution{IISc Bengaluru.}
  \country{Bengaluru, India}
  }

\author{Anshika Singh}
\email{anshika.singh@walmart.com}
\affiliation{%
  \institution{Walmart Inc.}
  \country{Bengaluru, India}
  }
    
\author{Priyanka Bhatt}
\email{priyanka.bhatt@walmart.com}
\affiliation{%
  \institution{Walmart Inc.}
  \country{Bengaluru, India}
  }

\begin{document}

\setcitestyle{authoryear,open={},close={}}

\begin{abstract}
Customer service chatbots are conversational systems aimed at addressing customer queries, often by directing them to automated workflows. A crucial aspect of this process is the classification of the customer's intent. Presently, most intent classification models for customer care utilise only customer query for intent prediction. This may result in low-accuracy models, which cannot handle ambiguous queries. An ambiguous query like ``I didn't receive my package'' could indicate a delayed order, or an order that was delivered but the customer failed to receive it. Resolution of each of these scenarios requires the execution of very different sequence of steps. Utilizing additional information, such as the customer's order delivery status, in the right manner can help identify the intent for such ambiguous queries. In this paper, we have introduced a context-aware NLU model that incorporates both, the customer query and contextual information from the customer's order status for predicting customer intent. A novel selective attention module is used to extract relevant context features. We have also proposed a multi-task learning paradigm for the effective utilization of different label types available in our training data. Our suggested method, Multi-Task Learning Contextual NLU with Selective Attention Weighted Context (MTL-CNLU-SAWC), yields a 4.8\% increase in top 2 accuracy score over the baseline model which only uses user queries, and a 3.5\% improvement over existing state-of-the-art models that combine query and context. 
We have deployed our model to production for Walmart's customer care domain. Accurate intent prediction through MTL-CNLU-SAWC helps to better direct customers to automated workflows, thereby significantly reducing escalations to human agents, leading to almost a million dollars in yearly savings for the company.

% Higher accuracy results in enhanced customer experience, reducing escalation rates to human agents and thus reducing cost. The model has been deployed to production and the results from A/B test results, it is expected to reduce escalations to human agents by almost 15\% leading to millions of dollars in yearly savings for the company.
%A/B test results indicate significant cost savings annually from the model.
\end{abstract}

\maketitle
\pagestyle{plain}



\section{Introduction}
Conversational agents have become integral to modern life, aiding in tasks such as playing music, ordering food, booking flights [\citep{8576921}] and banking [\citep{10128582}]. They have also notably improved customer care by efficiently resolving queries, boosting customer satisfaction and saving companies millions. A key component of these agents is Natural Language Understanding (NLU), which is responsible for deciphering the semantic meaning or intent from a user input. Our work in this paper is focused on the intent classification task and is primarily geared towards the customer care sector.

Over the past few years, the idea of fine-tuning a generalised pre-trained model for downstream tasks, introduced by \citep{howard2018universal}, has become very popular. Many applications employ transformers [\citep{vaswani2017attention,wolf2020transformers}] as the pre-trained model, achieving excellent results for various NLU tasks, including intent classification [\citep{rafiepour2023ctran,chen2019bert,lorenzo2021intformer}]. In the customer care domain, \citep{wang2021cs} and \citep{senese2020mtsi} have fine-tuned transformer models like BERT [\citep{devlin2018bert}] for the purpose of intent classification. However, most of these models rely solely on user utterances, which can be problematic when utterances are vague or lack sufficient detail, as often found in customer care settings. In such cases, providing additional contextual cues to the model can greatly improve its intent classification capability. Table \ref{table0} provides examples of such vague utterances where context can be beneficial.

One study that utilizes context to classify user intents is \citep{gupta2019context}. In this work, the model is trained using multiple signals, such as previous intents, slots and utterances over a variable context window. Here, context refers to the additional explicit input provided by the user. Another study by \citep{lu2019goal} leverages user profile information, like membership status, as context to generate responses. However, neither of these incorporate information from user's order history as context. In our work, we use the user's previous transactions to calculate features, such as order delivery status, order cancellation status etc. and combine them with the user query to predict intents.

Our context features are either binary, categorical or numerical, while the user query is textual. Hence, a mechanism that effectively combines these different data modalities is crucial. One notable work on this area by \citep{kaas2020team} uses stacked generalization to merge neural BERT representations of text with hand-crafted features. Another research by \citep{gu2021package} evaluates various feature combination techniques to combine data from these modalities and benchmarks performance on different datasets and tasks. Top performing architectures are Unimodal (feeds text along with tokenized numerical/categorical data directly to the transformer), MLP+Concat (passes numerical/categorical data through MLP block before concatenation with text embedding), Gating [\citep{rahman2020integrating}], and the weighted sum of feature vectors from different modalities.
In our work, we use an attention mechanism to dynamically attend to each context feature based on the user query and context, which is then concatenated with the query embedding and passed through a transformer architecture to predict intents.

For deep learning models to perform optimally, high-quality labeled data is essential. However, as previously noted, in the customer care domain, we often encounter ambiguous user queries. Consequently, for many examples, labels derived solely from user utterances (utterance labels) may not align with the actual user intent. Some prior studies addressing the issue of learning from noisy labels in real-world scenarios were performed by \citep{wei2021learning}, \citep{frenay2013classification} and \citep{huang2023twin}. To tackle this challenge, we introduced conversation labels in addition to utterance labels. We developed an architecture that utilizes the multi-task learning framework to effectively leverage both. The key contributions of this work can be summarized as follows:
\begin{itemize}
    \item Crafting features from the user's raw transaction data to streamline the model's learning process
    \item A novel attention mechanism that enables the model to dynamically focus on context features based on several factors
    \item A labeling approach that considers both, the users' explicit and latent intents
    \item A multi-task learning architecture that effectively leverages the above labeling strategy
\end{itemize}

\begin{table}
  \scalebox{0.8}[0.8]{
  \begin{tabular}{p{0.3\linewidth} | p{0.82\linewidth}}
    \toprule
    \textbf{Utterance}&\textbf{Possibilities}\\
    \midrule
    order cancelled & unclear whether user wants an order cancelled or is complaining about an order cancelled by store\\
    \midrule
    my order & will require different resolution depending on whether the user is talking about a delayed order or missing order\\
    \midrule
    not received & not clear whether the user is talking about order or refund\\
  \bottomrule
    \end{tabular}}
    \caption{\textmd{Vague utterances where context can help}}
\label{table0}
\end{table}

\section{Methodology}
\subsection{Constructing context features from customer's transaction data}\label{feature-creation}
Customer transaction data includes order-level details like order placement time, item count, store number, delivery fulfillment type etc. as well as item-level details like delivery status, cancellations, and refund requests for individual items. We combined item-level features corresponding to an order to create features such as ``number of items delivered'', ``time difference between last delivered item and user chat'', ``time difference between last shipped item and user chat'' and more. Further, features like ``are any items left to be delivered'', ``are any items left to be shipped'' and so on are created using these new features. For example, the feature ``are any items left to be delivered'' is created by comparing the number of ordered and delivered items for an order. The complete list of context features used in the model can be found in Table \ref{handcrafted-features}. 
\begin{table}[h]
  \scalebox{0.78}{
  \begin{tabular}{p{0.9\linewidth} | p{0.3\linewidth}}
    \toprule
    \textbf{Feature}&\textbf{Feature type}\\
    \midrule
    time in hours since order placement & numerical\\
    time in hours since last delivered/shipped item of order & numerical\\
    order fulfillment type & categorical\\
    order cancellation reason & categorical\\
    store indicator & categorical\\
    are any items left to be delivered/shipped & binary\\
    are all items left to be delivered/shipped & binary\\
    are any items past expected delivery time & binary\\
    are all items past expected delivery time & binary\\
    were any items substituted by store & binary\\  
    were any items cancelled by store/customer/other & binary\\  
    were all items cancelled by store/customer/other & binary\\ 
    was return for any item initiated by customer & binary\\  
    was return for all items initiated by customer & binary\\  
  \bottomrule
\end{tabular}}
\caption{\textmd{Examples of handcrafted features used. `/' sign is used to indicate 2 separate features. For example, time since last delivered item and time since last shipped item are two separate features}}
\label{handcrafted-features}
\end{table}

%For all the architectures discussed in the subsequent sections, we have utilized BERT [\citep{devlin2018bert}] as the encoder to obtain embedding from textual data.

\subsection{Baseline Model}
\begin{figure}
  \centering
  \includegraphics[width=1\linewidth]{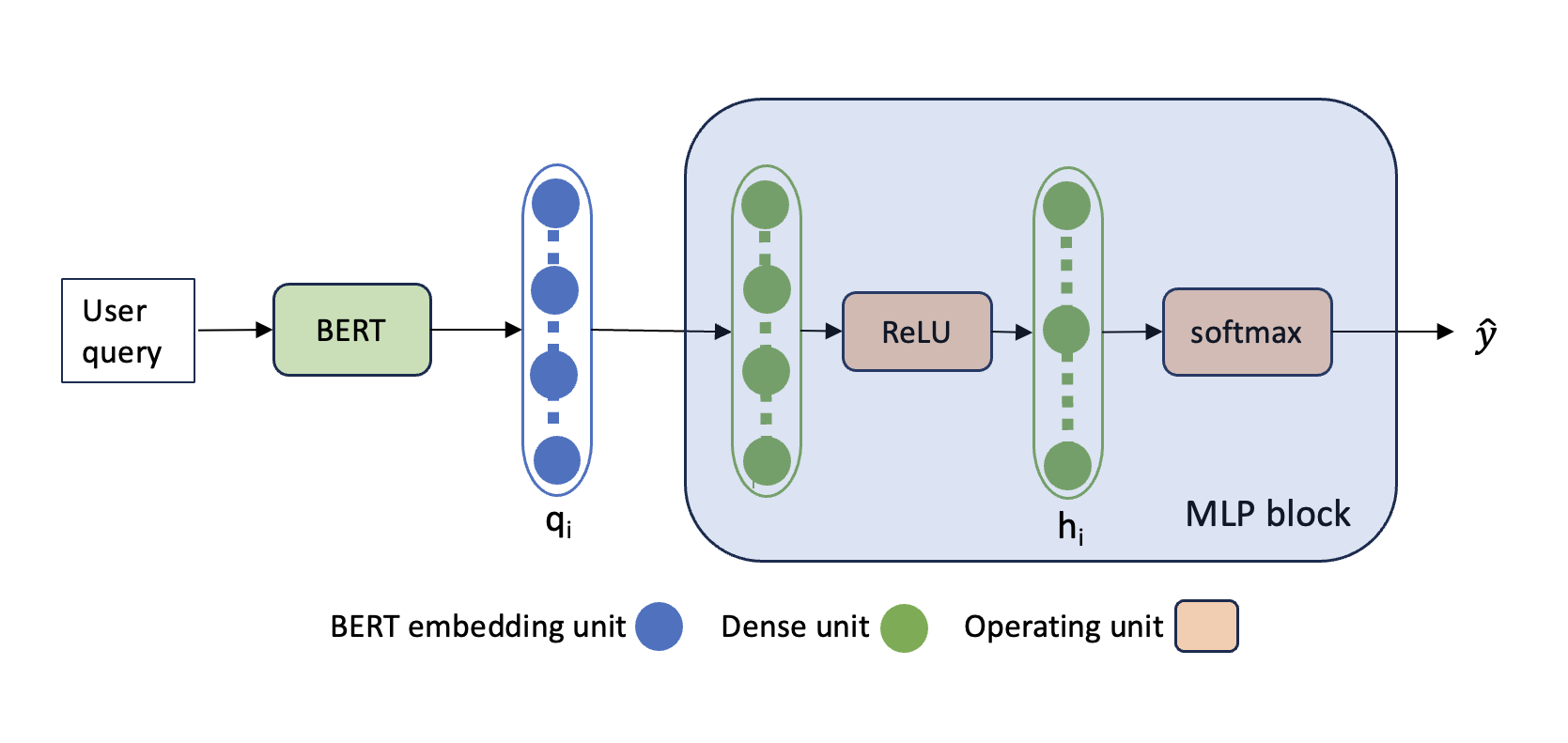}
  \caption{Baseline Model - \textmd{User query passed through pre-trained BERT model and subsequent MLP block}}
  \label{hnlu_fig}
\end{figure}

In our baseline model, the user utterance is passed through a pre-trained BERT model and the resulting embedding is fed into a Multi-Layer Perceptron (MLP) block with two hidden layers, that use ReLU [\citep{fukushima1975cognitron}] and softmax [\citep{bridle1989training}] activation functions respectively. The architecture is shown in Figure \ref{hnlu_fig}. Cross entropy loss, computed using the softmax layer's output and ground truth label, is used to train the model. For all the architectures discussed in the subsequent sections also, we have utilized BERT as the encoder to obtain embedding from textual data. More details of model and loss function can be found in section \ref{baseline_loss} of Appendix .

\subsection{Architecture to combine user utterance and context}
The categorical features are converted to fixed size embeddings via the process of Category Encoding which involves the use of a learnable embedding matrix for each feature. Post this, the categorical feature embeddings and other numerical and binary features undergo a pre-processing step, which involves min-max normalization and missing value imputation. This forms the context vector which is then combined with the query embedding. The combined embedding is fed through an MLP block. Among the numerous techniques to combine query and context, the most straightforward approach is to concatenate the two. However, the subsequent MLP layers often struggle to attend to relevant information in the context vector for a given query. Consequently, the Concat [\citep{gu2021package}] model's performance is unsatisfactory.

\subsubsection{\textbf{Concat with Attention weighted Context (CAWC)}}
\label{attention}
\begin{figure}
\centering
\includegraphics[height=0.47\linewidth,width=1\linewidth]{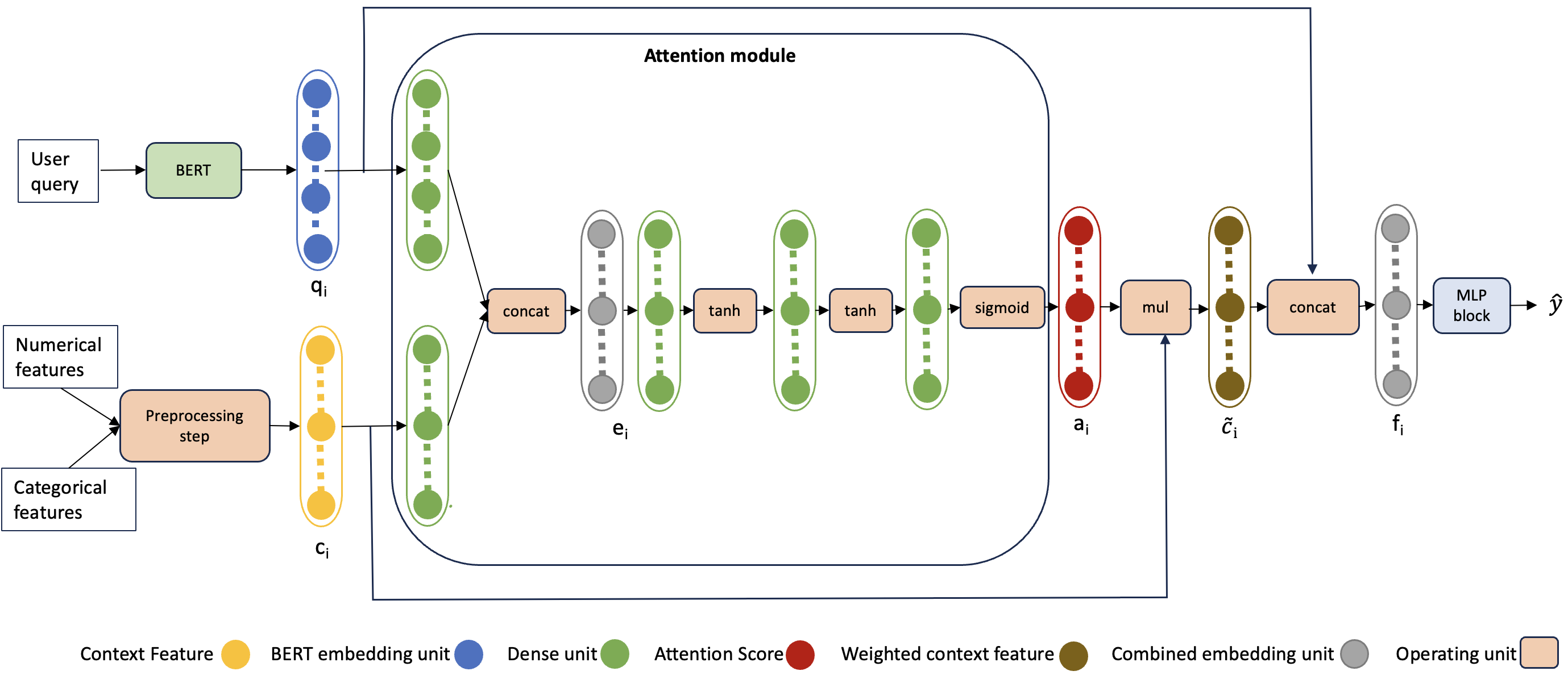}
  \captionsetup{width=1\linewidth}
  \caption{\textmd{Concat with Attention Weighted Context}}
  \label{fig:attention_fig}
\end{figure}

We utilize an attention-based feature weight generation mechanism in which attention weights are computed for each of the context features, based on both the context and query embeddings. This approach enables the model to concentrate on relevant features, significantly mitigating the issues associated with the Concat model. The attention module receives $q_i$ (query vector) and $c_i$ (context vector) and produces an attention vector $a_i$ with the same length as $c_i$. The weighted context vector $\tilde{c}_i$ is then obtained by performing element-wise multiplication of $a_i$ and $c_i$. Finally, the weighted context vector is concatenated with the query embedding, which is then input to an MLP block as before. 
\begin{equation}
     a_i = \text{AttentionModule}(q_i, c_i), \;\;\; \hat{c}_i = a_i \odot c_i
\end{equation}
where $\odot$ represents element-wise multiplication.\\
\newline
\textbf{Attention Module:}\label{attention-module}
 In the attention module, both the query and context vectors are passed through linear layers represented by $W_q$ and $W_c$, respectively, and subsequently concatenated to form a combined vector denoted as $e_i$. The vector $e_i$ is then fed through two more linear layers, $W_{l1}$ and $W_{l2}$, with the number of neurons roughly halved in each hidden layer. The tanh activation function is applied after each linear layer. Following this, the resulting vector is passed through another linear layer, denoted by $W_{l3}$, whose output vector length is equal to that of the context vector. The sigmoid activation function, $\sigma$, is then applied to adjust each value to a scale between 0 and 1. The resulting attention vector is represented by $a_i$.

\begin{equation}
     e_i = \text{concat}(W_q q_i, W_c c_i)
\end{equation}
\begin{equation}
     a_i = \sigma(W_{l3} \tanh(W_{l2} \tanh(W_{l1} e_i)))
\end{equation}
where $\sigma$ represents the sigmoid function.
\\
The complete architecture is shown in Figure \ref{fig:attention_fig}.

\subsection{Labelling Strategy}\label{labelling strategy}
We have discussed how using context can help predict more accurate intents, especially with ambiguous user utterances. Similarly, having context information also aids in the proper annotation of data. In our case, labeling examples solely on user utterances might result in incorrect labels when the user query is vague. Nevertheless, the accurate approach involves examining the user's query along with all its contextual features for labeling the examples. This method is extremely time-consuming and not scalable for large datasets. To tackle this, we use examples from our dataset where the query could not be resolved by chatbot and an agent had to step in. The subsequent user-agent conversation serves as a proxy for the user's context, and we use this information for labelling the data. This is effective as it often contains the same context information available in the user's transaction history. Therefore, we have two types of labels for each example:
\begin{itemize}
\item \textbf{Utterance Label:} tagged based solely on the user's utterance and is intended to capture the user's explicit intent
\item \textbf{Conversation Label:} tagged based on the entire user-agent conversation and captures the user's latent intent.
\end{itemize}
It is important to note that during inference, we only have access to user query and context. Agent conversations are used solely for obtaining conversation labels for training data, which serve as a substitute for labels tagged based on context. The challenge is to effectively use both label types. Utterance labels may not capture the user's latent intent, while conversation labels might miss the explicit user intent. For example, for the user input ``contact customer care'', the utterance label is ``agent contact''. However, the conversation label might be ``refund'', tagged based on further user-agent conversation. For such examples, the model should be able to predict both. To equip our model with the capability of efficiently predicting both intents, we use a multi-task learning framework for training our model.

\begin{figure}[t]
  \centering
  \includegraphics[height=0.5\linewidth,width=1\linewidth]{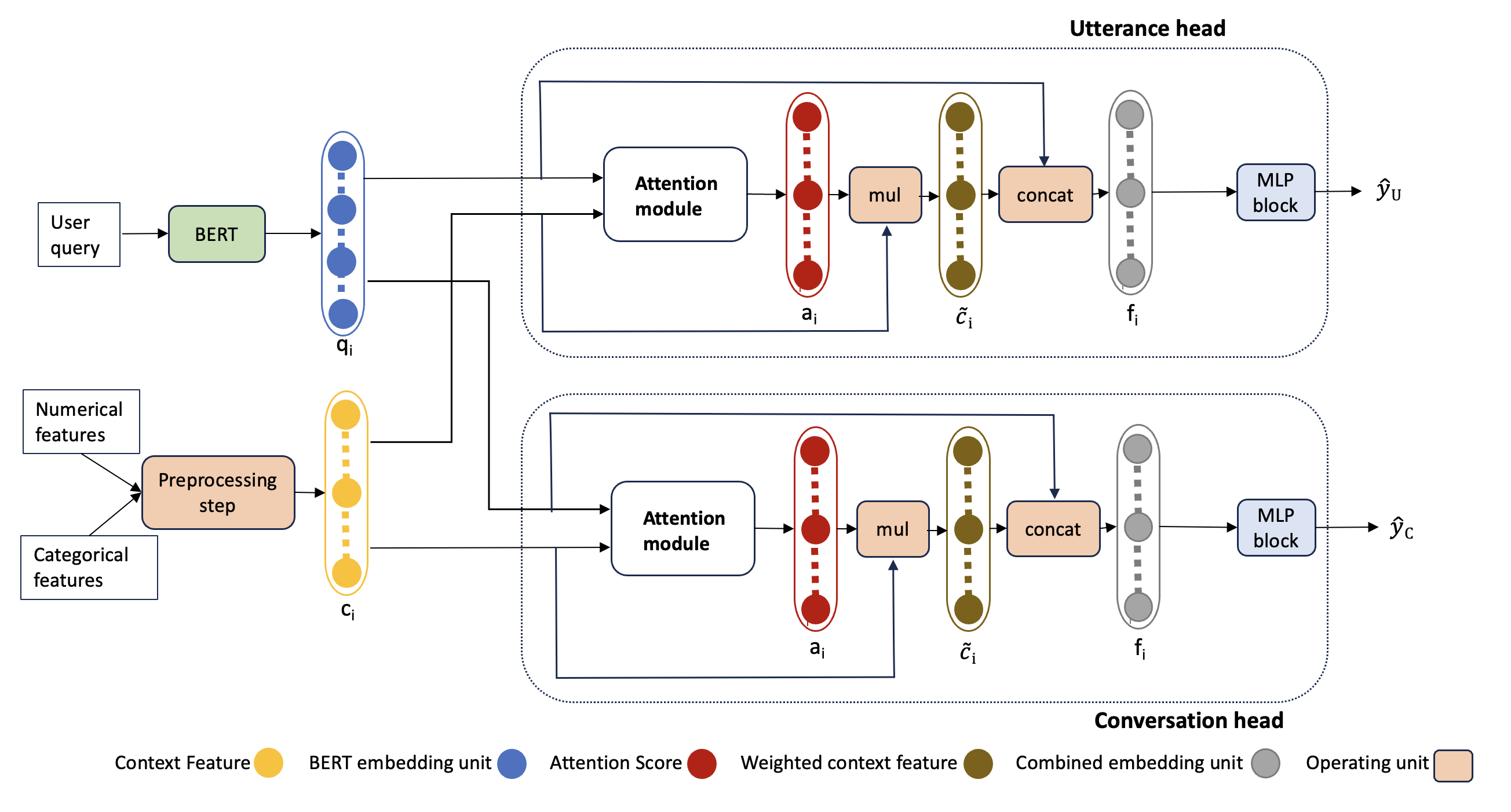}
  \caption{MTL-CNLU: \textmd{Dual headed model trained using multi-task learning with utterance and conversation labels}}
  \label{MTL-CNLU}
\end{figure}

\begin{figure*}[t]
  \centering  
  \includegraphics[height=0.37\linewidth,width=0.75\linewidth]{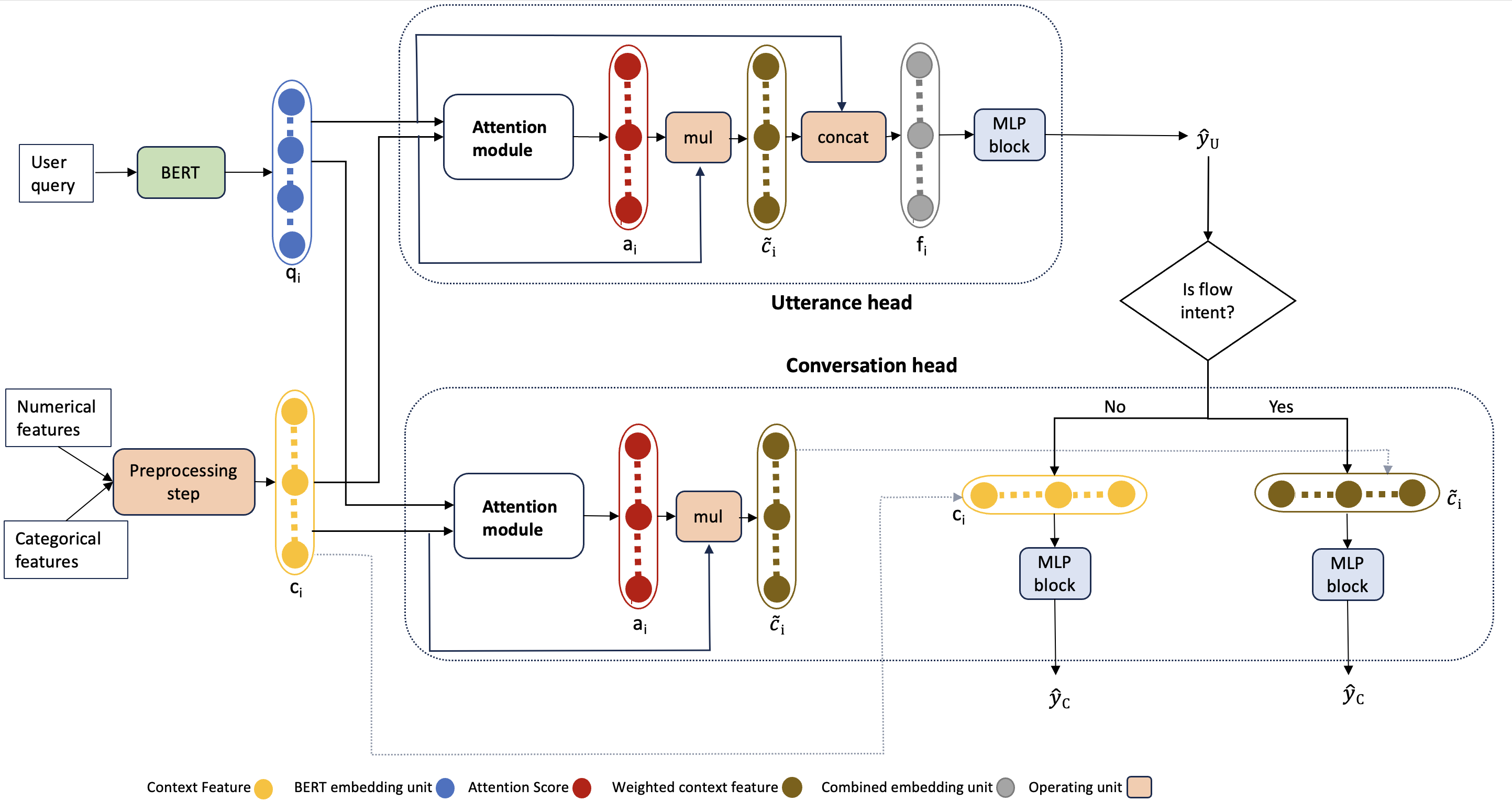}
  \captionsetup{width=1\linewidth}
  \caption{MTL-CNLU-SAWC: \textmd{Attention weighted context vector used to predict conversation intents when user query corresponds to flow intent, otherwise context features used directly}}
  \label{MTL-CNLU-SAWC}
\end{figure*}

\subsection{Multi Task learning paradigm for Contextual NLU (MTL-CNLU)} \label{MTL-CNLU_section}
Multi-Task Learning (MTL) [\citep{zhang2021survey,zhang2022survey}] is a training framework employed to exploit valuable information within multiple related tasks, thereby enhancing the generalization performance across all tasks. In our case, we consider two tasks: utterance intent (or explicit intent) classification and conversation intent (or latent intent) classification, using utterance and conversation labels as their respective ground truths. As noted in \citep{ruder2017overview} most MTL models share the initial layers among tasks. In our scenario, the backbone BERT module is shared between the utterance head (accountable for predicting utterance intent) and the conversation head (accountable for predicting conversation intent). The parameters of the backbone BERT module, utterance head, and conversation head are denoted by $\phi_1$, $\phi_{2U}$, and $\phi_{2C}$, respectively. The combined loss from the two heads is utilized to jointly update all three parameter sets and is represented by:
\begin{equation}
{\mathcal{L}}_{combined} = {\mathcal{L}}(D;\phi_1,\phi_{2U},\phi_{2C})
\end{equation}
The combined loss is a weighted sum of the cross-entropy losses from the two heads.
\begin{equation}
{\mathcal{L}}_{combined} = {\mathcal{L}}_{ce}(Y_u,\hat{Y}_u) + \lambda {\mathcal{L}}_{ce}(Y_c,\hat{Y}_c) 
\end{equation}
where $\lambda$ is a hyperparameter used to balance the two losses. $Y_u$, $Y_c$, $\hat{Y}_u$, and $\hat{Y}_c$ represent the utterance labels, conversation labels, predicted utterance intents and conversation intents respectively.

\subsection{Architectures for MTL-CNLU}

\subsubsection{\textbf{MTL-CNLU}}
The pre-trained BERT module is the only component shared by both the utterance and conversation heads. Each head possesses its own query-context combining module, which maintains the same architecture as CAWC (introduced in section \ref{attention}). The architecture is shown in Figure \ref{MTL-CNLU}. The predicted intents from the utterance and conversation heads are represented by $\hat{y}_u$ and $\hat{y}_c$, respectively. We also experimented with an architecture where the query-context combining module was shared by both heads. This resulted in poorer performance and was therefore discarded. More details on this can be found in section \ref{alternate_arch} of Appendix.

\subsubsection{\textbf{MTL-CNLU with Selective Attention Weighted Context (MTL-CNLU-SAWC)}}\label{only-weighted-context} 
The concatenation of the query and weighted context vector for predicting intents works effectively for the utterance head. However, due to the context vector's sparse nature and its comparatively shorter length (context vector is of length 93, made up of 18 binary/numerical features and the rest from categorical feature embeddings, while the query vector is of length 768), concatenating the two causes the model to focus more on the query than the context. This results in suboptimal outcomes for the conversation head. For instance, our training data contained an utterance ``hello there'' with ``greet'' and ``why order was cancelled'' as its explicit and latent intents respectively. However, since the model learns to place a greater emphasis on user query for its prediction of latent intent also, during inference, when a user with a latent intent of tracking their order status queries ``hello there'', the model predicts ``greet'' and ``why order was cancelled'' as the explicit and latent intents instead of ``greet'' and ``where is my order''. Owing to findings like these, and since the conversation head's primary objective is to predict the user's latent intent, the previous architecture was modified to remove the query-context combining module from the conversation head. The weighted context vector is fed directly into the MLP block to predict latent intents. This change alone enhances the model's ability to predict latent intents. Another way in which MTL-CNLU-SAWC improved upon MTL-CNLU was the effective use of flow and non-flow intents. Each predicted intent is one of the two types:
\begin{itemize}
    \item \textbf{Flow intent:} Intents associated with a defined flow. When a user selects a flow intent, they can follow a series of predefined steps to resolve their query. Examples of flow intents include ``where is my order'', ``why order was cancelled'' and ``where is my refund''.
    \item \textbf{Non-flow intent:} Intents that do not have an associated flow. These include intents like ``agent contact'', ``greet'', ``affirmative'' etc.
\end{itemize}
As detailed in Section \ref{attention-module}, the attention weights for the context features are derived using both context and query embedding. However, for utterances such as ``contact customer care'', the query contains no relevant information regarding the user's latent intent. Therefore the query embedding should ideally not influence the latent intent prediction. In fact, for all utterances where the explicit intent corresponds to a non-flow intent, the latent intent prediction should depend solely on the context vector. To accomplish this, we modify the architecture so that the context vector $c_i$ is element-wise multiplied with the attention vector $a_i$ only when the utterance head predicts a flow intent. Otherwise, $c_i$ is directly fed as input to conversation head. Architecture details can be found in Figure \ref{MTL-CNLU-SAWC}.
\begin{algorithm}
\caption{Selectively apply attention to context vector}
\label{alg:sel-att}
\begin{algorithmic}
\If{$\hat{y}_u \in$ Flow Intents List} 
    \State $\tilde{c} \gets a_i*c_i$
\Else
    \State $\tilde{c} \gets c_i$
\EndIf 
\end{algorithmic}
\end{algorithm}

\section{Experimental Setup}
\subsection{Dataset}
The experimental data is sourced from the Walmart's customer care domain. Each observation consists of a user utterance, context features, and two labels: utterance and conversation labels. Since the data is Walmart's internal user data, it has not been disclosed. However, some characteristics of the data have been provided in Table \ref{dataset-stats}. We could not benchmark the performance of our model on open source data because, to the best of our knowledge there is no open source e-commerce intent classification dataset that also contains customer order information. Our approach however is generic enough to be applied to the customer care domain of any e-commerce firm because the context features we construct are not specific to our dataset and can be easily created if a customer's order information is available. 
\begin{table}[h]
  \scalebox{0.8}{
  \begin{tabular}{p{0.99\linewidth} | p{0.15\linewidth}}
    \toprule
    \textbf{Dataset detail}& \textbf{Statistics}\\
    \midrule
    \#utterances in training  & 100K\\
    \#utterances in validation  & 2.5K\\
    \#utterances in test  & 2.5K\\
    \#context features & 21\\
    \% of e.g. with context & 70\%\\
    \% of e.g. with distinct utterance \& conv. label & 45\%\\
    \#intents covered by utterance labels & 59\\
    \#intents covered by conversation labels & 35\\
  \bottomrule
\end{tabular}}
\caption{\textmd{Dataset statistics}}
\label{dataset-stats}
\end{table}

\begin{table*}[t] 
  \scalebox{0.90}{
  \begin{tabular}{c c c c c c}
    \toprule
    \textbf{Architecture}&\multicolumn{2}{c}{\textbf{Utterance Intent}}&\multicolumn{2}{c}{\textbf{Conversation Intent}}& \multicolumn{1}{c}{\textbf{Top 2 Score(\%)}}\\
    & \textbf{Micro F1(\%)}&\textbf{Macro F1(\%)}&\textbf{Micro F1(\%)}&\textbf{Macro F1(\%)}&\\
    \midrule
    Text only (baseline)               & 78.65 & 75.80& - & - & 86.12\\
    Concat [\citep{gu2021package}]     & 80.14 & 77.28& - & - & 87.34\\
    MLP + Concat [\citep{gu2021package}] & 80.28 & 77.66& - & - & 87.23\\
    Unimodal [\citep{gu2021package}]     & 79.66 & 76.01& - & - & 86.14\\
    Gating [\citep{gu2021package,rahman2020integrating}]      & 80.45 & 77.42& - & - & 87.41\\
    Weighted Sum [\citep{gu2021package}] & 80.12 & 77.37& - & - & 86.98\\
    \midrule
    CAWC (Ours) & 81.5 & 78.71& - & - & 88.38\\
    MTL-CNLU (Ours)     & 81.65 & 78.80& 38.65 & 37.80& 89.90\\
    MTL-CNLU with shared query-context combining module (Ours)     & 81.48 & 78.62& 34.65 & 32.76& 85.90\\
    \textbf{MTL-CNLU-SAWC} (Ours)  & \textbf{81.96} & \textbf{79.05}& \textbf{42.03} & \textbf{39.56}& \textbf{90.92}\\
    \midrule
    GPT4 (one shot) & 63.0 & 58.62 & 30.44 & 27.61 & 74.33\\
    RAG with GPT4     & 71.88 & 65.80 & 32.62 & 30.75& 83.98\\
  \bottomrule
\end{tabular}}
\caption{\textmd{The table compares performances of various models. The top half lists results from the baseline model (text only) and various SOTA models in \citep{gu2021package}. Our models fill the middle section, followed by LLM models. The first 7 models, with single-headed architectures, could not be compared with the conversation label.}}
\label{quantitative}
\end{table*}

\begin{table*}[t]  
  \scalebox{0.88}{
  \begin{tabular}{|p{0.2cm}|p{2.3cm}|p{2.8cm}|p{1.5cm}|p{2cm}|p{1.2cm}|p{1cm}|p{2.35cm}|p{1cm}|p{1cm}|p{1cm}|p{1.2cm}|}
   \hline
   &\textbf{Utterance}&\textbf{Context Information}&\textbf{utterance label}&\textbf{conversation label}&\textbf{Baseline}&\textbf{concat}&\textbf{MLP+concat/ Unimodal/ Gating/ Weighted Sum}&\textbf{CAWC}&\textbf{MTL-CNLU}&\textbf{MTL-CNLU-SAWC} \\
   \hline
   1 & my order is late & items are overdue & order late& order late& \doublecheck &\doublecheck &  \doublecheck& \doublecheck& \doublecheck& \doublecheck\\
   \hline
   2 & cancel my order & items are yet to be delivered & cancel order& cancel order& \doublecheck &\doublecheck & \doublecheck& \doublecheck& \doublecheck& \doublecheck\\
    \hline
    3 & order cancelled & store cancelled order & why order cancelled &why order cancelled& \cross  & \doublecheck& \doublecheck& \doublecheck& \doublecheck& \doublecheck \\
    \hline
    4 & when will i receive the items & all items overdue & order late&order late& \cross &\cross &  \cross& \doublecheck& \doublecheck& \doublecheck\\
    \hline
    5 & order help & items were missing from order & where is my order & missing items& \checked &\checked &\checked& \checked& \doublecheck& \doublecheck\\
    \hline
    6 & hello & all items overdue &greet &order late &\checked & \checked& \checked& \checked& \checked& \doublecheck\\
  \hline
  \end{tabular}}
\caption{\textmd{Qualitative analysis of different architectures. Double check mark (\doublecheck)
indicates that model outputs cover both the labels, single check mark (\checked) indicates that model outputs cover only one of the two labels and cross (\cross) indicates none of the labels are covered.}}
\label{qualitative}
\end{table*}
\subsection{Evaluation metrics}\label{evaluation-metrics}
\subsubsection{\textbf{Micro-F1 and Macro-F1 scores}}
The top 1 intent from utterance head and top 1 intent from conversation head are compared to the utterance and conversation labels respectively. Micro-F1 and Macro-F1 scores are used as performance metrics for both.

\subsubsection{\textbf{Top 2 score}}\label{top2_subsect}
Since we want our models to accurately predict the user's explicit as well as latent intent, it is important that we also formulate a metric that jointly accounts for both. We call this metric the top 2 score. To calculate the top 2 score for a model, we first obtain the top 2 intents predicted by it. For non-MTL architectures with a single head, these intents are the two intents with highest confidence scores. For MTL models, they comprise of the top intent from the utterance head and the top intent from the conversation head. The top 2 score measures how well these intents align with the utterance and conversation labels. The logic to calculate top 2 score can be found in Algorithm \ref{alg:top2}.
 
\begin{algorithm}
\caption{Top 2 score calculation}
\label{alg:top2}
\begin{algorithmic}
    \State Let $y_u$, $y_c$, $\hat{y}_1$ and $\hat{y}_2$ denote the utterance label, conversation label, 1st predicted intent and 2nd predicted intent respectively.
    \If {$y_u = y_c$} 
        \If {$\hat{y}_1 = y_u$  or  $\hat{y}_2 = y_u$}
            \State score $\gets$ 1
        \Else
            \State score $\gets$ 0
        \EndIf
    \Else    
        \If {$\hat{y}_1 \in  \{y_u,y_c\}$ and $ \hat{y}_2 \in \{y_u,y_c\}$}
            \State score $\gets$ 1
        \ElsIf {$\hat{y}_1 \in \{y_u,y_c\}$ or $ \hat{y}_2 \in \{y_u,y_c\}$}
            \State score $\gets$ 0.5
        \Else
            \State score $\gets$ 0
        \EndIf 
    \EndIf 
\end{algorithmic}
\end{algorithm}

\section{Results}
In Table \ref{quantitative}, the evaluation metrics from Section \ref{evaluation-metrics} are used to compare all the proposed architectures of this paper, as well as those studied by \citep{gu2021package}. Additionally, we have also benchmarked the performance of popular LLMs for our customer intent classification task, where both the customer query and context were fed to the LLM as input. To this end, our experiments involved few-shot prompting of GPT4 [\citep{openai2023gpt4}], and Retrieval Augmented Generation(RAG) [\citep{lewis2020retrieval}] with GPT4. More information on these experiments is available in section \ref{llm-exp}.

The results show that MTL-CNLU-SAWC outperforms all other models across all metrics. Among the single-headed architectures, CAWC performs the best, due to the successful query and context information integration via its attention module. MTL-CNLU-based architectures enhance the top 2 accuracy scores by better predicting the user's latent intent. MTL-CNLU-SAWC further improves latent intent prediction by selectively using information from the user query. 

Table \ref{qualitative} provides a qualitative analysis of different architectures' outputs using varied examples. In examples 1 and 2, the utterance is clear and all models are able to predict intents accurately. The baseline model incorrectly identifies ``cancel order'' as the intent in example 3. In example 4, the concat and \citep{gu2021package} architectures fail to recognize the overdue delivery status and incorrectly predict ``missing items'' as the intent. The attention-based architecture (CAWC), however, accurately identifies ``order late'' as the intent using context information. In examples 5 and 6, utterance and conversation labels differ. Only the MTL-CNLU models correctly predict both intents in example 5. In example 6, where the query lacks relevant information about user's latent intent, the MTL-CNLU model mispredicts the latent intent but the MTL-CNLU-SAWC model correctly predicts both intents. 

\subsection{Ablation Study}
As stated in section \ref{feature-creation}, we have derived several features from user transaction data and used them as context in our model. To evaluate the significance of the different order level, item level and handcrafted context features, we compared the model's performance with and without them. The findings from this analysis have been presented in Table \ref{feature-contribution}. Handcrafted features, such as ``are any items left to be delivered'' and ``were any items cancelled'' have the strongest influence on model accuracy. We utilized the CAWC architecture to make these comparisons.
\begin{table}[h]
  \scalebox{0.75}{
  \begin{tabular}{p{0.9\linewidth} | p{0.3\linewidth}}
    \toprule
    \textbf{Model trained on}&\textbf{top 1 accuracy (\%)}\\
    \midrule
    text only (baseline) & 78.65\\
    text + order level features  & 79.34\\
    text + item level features  & 78.91\\
    text + order level + item level features  & 79.42\\
    \textbf{text + order level + item level features + handcrafted features}& \textbf{81.48}\\
    text + order level + item level features + handcrafted features - \{``are any items left to be delivered''\} & 79.86\\
    text + order level + item level features + handcrafted features - \{``were any items cancelled''\} & 80.41\\
    
   \bottomrule
  \end{tabular}}
 \caption{\textmd{Assessment of the contribution of different context features towards model performance improvement}}
 \label{feature-contribution}
\end{table}

\section{Training details}
\label{train-details}
All model architectures were trained on the same dataset, differing only in the training methodology for MTL-CNLU models due to the use of conversation labels.
The BERT model, employed as an encoder in all architectures is a pre-trained distilled variant, Small BERT [\citep{turc2019}], with a hidden state dimension of 768. All models were trained using the AdamW [\citep{loshchilov2017decoupled}] optimizer with a learning rate of 0.0001. A dropout of 0.5 was applied to the layers in the MLP block. The first hidden layer in the MLP block utilized the ReLU activation function, while the tanh activation function was used after layers in the attention module. The choice of activation function was based on empirical results. In MTL-CNLU, conversation labels serve as stand-ins for context-based user labels and do not affect training without context. Consequently, the hyperparameter $\lambda$ is set to zero for training examples without context data. For the remaining examples, we experimented with values \{0.6, 0.8, 1, 1.2, 1.4\} for $\lambda$ and ultimately set it to 1. A batch size of 32 was employed during training. The Tensorflow library was used for implementation, and all models were trained on an Nvidia V100 GPU. Our final model, MTL-CNLU-SAWC, contains approximately 39 million trainable parameters and takes roughly 3 hours to train on an Nvidia V100 GPU.

\section{LLM experiments}
\label{llm-exp}
\subsection{Few-shot prompting GPT4}
In this experiment we feed the user query and context features in words to GPT 4 and instruct it to predict two intents one primarily based on the user query and the other based on context. The prompt also contains the list of explicit and latent intents to choose from and 10 randomly selected labelled examples from our training data.
\subsection{RAG with GPT4}
Here, we make use of the RAG paradigm to ensure that only examples relevant to the current query are part of the prompt to GPT4. Embeddings are computed from all (customer utterances, context) pairs in our training data and stored into a Milvus [\citep{wang2021milvus}] vector database collection, along with other information. The following fields are stored for each example in the database.
\begin{itemize}[noitemsep,topsep=1pt]
    \item User query + context in words
    \item An embedding for (User query + context)
    \item utterance label and conversation label
\end{itemize}
During inference, the joint embedding of the user query and context is calculated and 10 entries from the vector db whose embeddings have the highest cosine similarity with it are added to the prompt as examples. The rest of the prompt to GPT4 remains the same. As evidenced by the results, this method provides a significant boost to the GPT4 performance however, the performance still falls well short of our best performing model. For calculating embeddings, we use bge-base-en [\citep{xiao2023c}] embedder.

\section{Conclusion}
In this paper, we have presented an effective approach for combining user queries and context information for the intent classification task in the customer care domain. Our proposed model outperforms both the baseline model, which only uses the user query, as well as other existing state-of-the-art models that combine query and context. We experimented with different types of context features and created several features manually to simplify the learning process of the model. We addressed the problem of noisy labels by adding conversation labels to utterance labels and developed an effective multi-task learning approach to leverage both simultaneously. We also demonstrated that condition-based selective application of attention weights can further enhance model performance. Our proposed method is generic enough to be applicable to the customer care domain of any e-commerce organisation. All we need are customer queries, customer order information corresponding to the queries, and utterance and conversation labels. Generation of context vector from order information is a straight forward forward process and does not involve the creation of any feature that is specific to our dataset. For applications where conversation labels are not available, our CAWC model can serve as benchmark.

\section{Future Work}
In our current work, which focuses on the e-commerce customer care domain, we employ context information from the customer's transaction history in conjunction with the customer query to predict intents. However, a potential source of context information that remained untapped in this work is the customer's website/app interaction data. This could be particularly valuable if the customer had recently visited help pages before engaging in chatbot conversation. Another source of context information that we can make use of in future iterations of this work, is the conversations of the customer with human agent/chatbot from an earlier session.

\bibliographystyle{ACM-Reference-Format}
\bibliography{custom}

\appendix

\section{Details of the baseline model}
\label{baseline_loss}
Given dataset D = ($x_i$ , $y_i$) with N different classes and M examples, we fine-tune BERT and train the MLP block layers.
The output probability from the model can be represented as follows:
\begin{equation}
p(y|h_i) = \text{softmax}(h_i) \in R^N 
\end{equation}
where $h_i$ $\in$ $R^N$ is the output from the last layer of the MLP block before the application of softmax, for the i-th example $x_i$. The model parameters $\theta$ are trained on D with cross-entropy loss.
\begin{equation}
\theta^* = \operatorname*{argmin}_\theta{\mathcal{L}}_{ce}(D;\theta)
\end{equation}
Cross entropy loss is defined as: 
\begin{equation}
{\mathcal{L}}_{ce} = -\sum_{i=1}^M\sum_{c=1}^Ny_{i,c}\log(p_{i,c}) \label{ce}
\end{equation}
where $p_{i,c}$ is the predicted probability of the i-th example belonging to class c, and $y_{i,c} \in \{1,0\}$, depending on whether c is the true class for the i-th example or not. %$\hat{y}$ in Figure \ref{hnlu_fig} denotes the intent class with the maximum probability.

\begin{figure}[h]
  \centering
  \includegraphics[height=0.4\linewidth,width=1\linewidth]{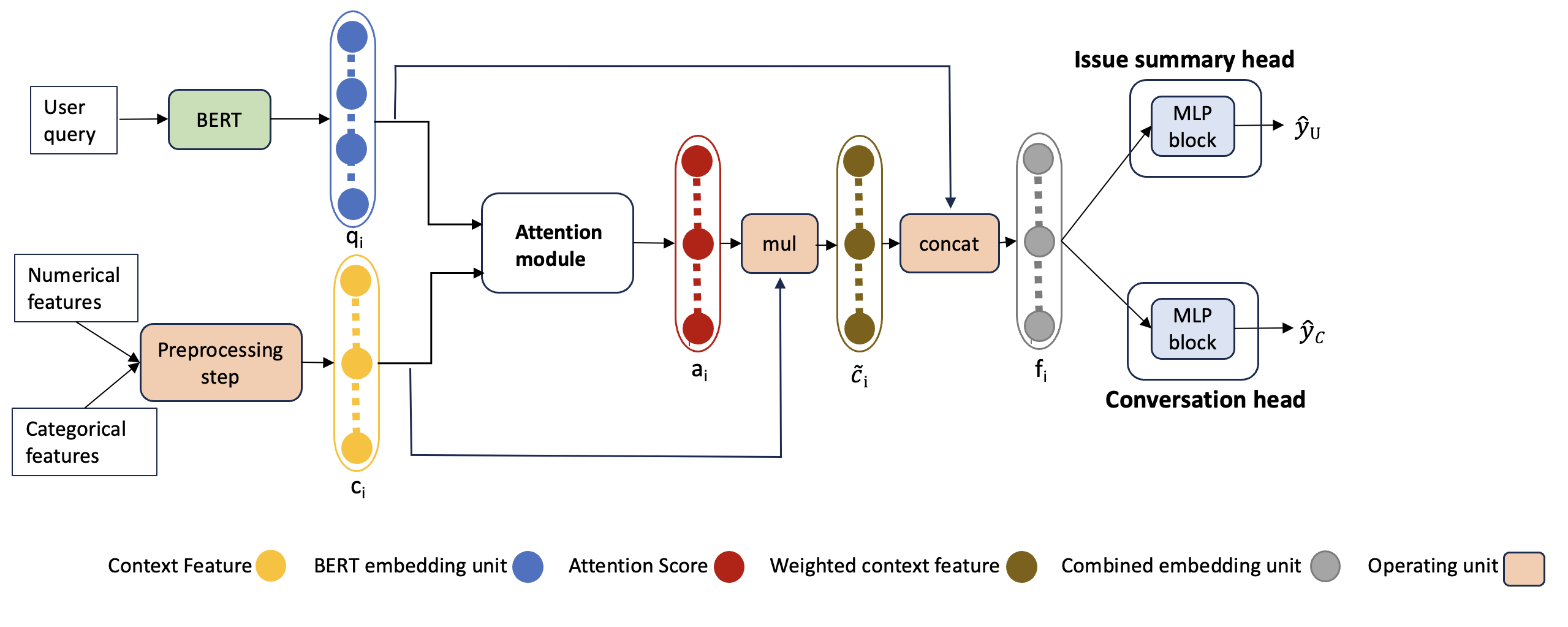}
  \captionsetup[figure]{font=small,skip=0pt}
  \caption{MTL-CNLU-shared: \textmd{MTL-CNLU with query-context combining module shared between the heads}}
  \setlength{\belowcaptionskip}{-35pt}
  \label{MTL-CNLU-shared}
\end{figure}
\begin{figure}[h]
  \centering
  \includegraphics[height=0.4\linewidth,width=1\linewidth]{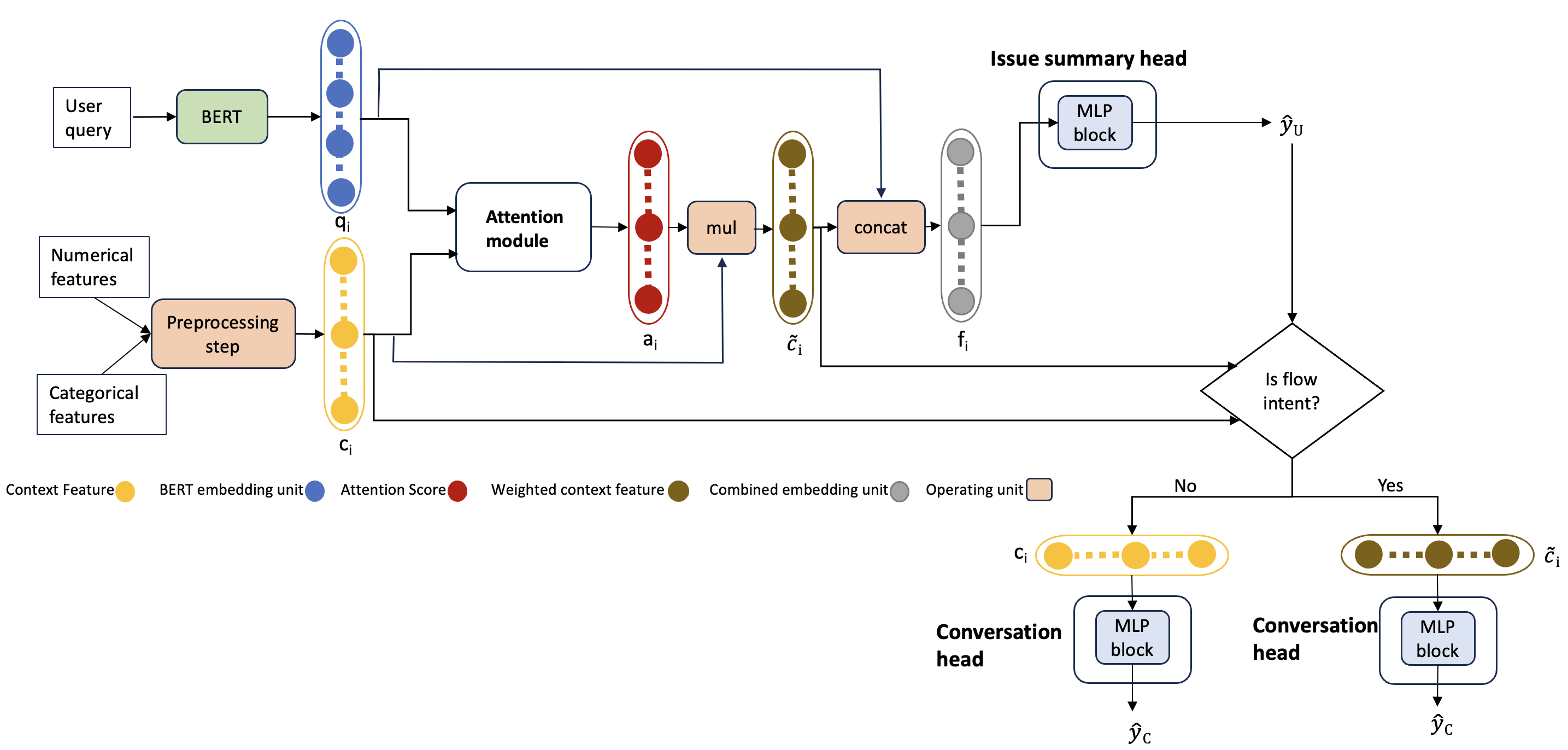}
  \caption{MTL-CNLU-SAWC-shared: \textmd{MTL-CNLU-SAWC with query-context combining module shared between the heads}}
  \label{MTL-CNLU-SAWC-shared}
\end{figure}

\begin{table*}[t]
\centering
  \scalebox{1}{
  \begin{tabular}{c c c c c c}
    \toprule
    \textbf{Architecture}&\multicolumn{2}{c}{\textbf{Utterance Intent}}&\multicolumn{2}{c}{\textbf{Conversation Intent}}& \multicolumn{1}{c}{\textbf{Top 2 Score (\%)}}\\
    & \textbf{Micro F1(\%)}&\textbf{Macro F1(\%)}&\textbf{Micro F1(\%)}&\textbf{Macro F1(\%)}&\\
    \midrule
    MTL-CNLU      & 81.65 & 78.80& 38.65 & 37.80& 89.90\\
    MTL-CNLU-SAWC & 81.96 & 79.05& 42.03 & 39.56& 90.92\\
    MTL-CNLU-shared      & 81.48 & 78.62& 34.65 & 32.80& 85.90\\
    MTL-CNLU-SAWC-shared & 81.62 & 78.81& 34.77 & 32.56& 87.18\\
  \bottomrule
\end{tabular}}
\captionsetup{width=0.8\linewidth}
\caption{\textmd{Results comparing performance of two sets of MTL-CNLU based architectures. Ones where both the query-context combining module and BERT module were shared between the heads, and the ones where only the BERT module was shared}}
\label{quantitative_shared}
\end{table*}

\section{Alternate architecture considerations for MTL-CNLU}\label{alternate_arch}
As outlined in section \ref{MTL-CNLU_section}, the core BERT module in all our MTL-CNLU-based architectures is shared by both the utterance and conversation heads. However, each head possesses its own query-context combining module. We experimented with using the same query-context combining module for both heads. This was carried out for both MTL-CNLU and MTL-CNLU-SAWC. The resulting new architectures are named MTL-CNLU-shared and MTL-CNLU-SAWC-shared respectively. Detailed architectures are available in figures \ref{MTL-CNLU-shared} and \ref{MTL-CNLU-SAWC-shared}. 
Each of these models demonstrated lower performance than their counterparts with separate query-context combining modules, as shown in Table \ref{quantitative_shared}.

\end{document}